\documentclass[lettersize,journal]{IEEEtran}
\usepackage{amsmath,amsfonts}
\usepackage{algorithmic}
\usepackage{algorithm}
\usepackage{array}
\usepackage[caption=false,font=normalsize,labelfont=sf,textfont=sf]{subfig}
\usepackage{textcomp}
\usepackage{stfloats}
\usepackage{url}
\usepackage{verbatim}
\usepackage{graphicx}
\usepackage{cite}
\usepackage{bbding}
\usepackage{amssymb}
\usepackage{booktabs}
\usepackage{multirow}
\usepackage[
    colorlinks=true,      
    citecolor=blue,       
    linkcolor=blue,       
    urlcolor=blue,        
    pdfstartview=FitH     
]{hyperref}

\def\BibTeX{{\rm B\kern-.05em{\sc i\kern-.025em b}\kern-.08em
    T\kern-.1667em\lower.7ex\hbox{E}\kern-.125emX}}
\usepackage{balance}

\begin{document}
 
\title{Calibrating Undisciplined Over-Smoothing in Transformer for Weakly Supervised Semantic Segmentation}

\author{

Lechao Cheng\thanks{Manuscript created March, 2025. (\textit{Corresponding Authors: Meng Wang, eric.mengwang@gmail.com})}, Zerun Liu, Jingxuan He\thanks{Lechao Cheng, Zerun Liu, Jingxuan He are with the School of Computer and Information Science, Hefei University of Technology, Hefei 230000, China (emails: 22021025@zju.edu.cn, 2112012095@zjut.edu.cn and chenglc@hfut.edu.cn).}, Chaowei Fang\thanks{Chaowei Fang is with the Key Laboratory of Intelligent Perception and Image Understanding of the Ministry of Education, School of Artificial Intelligence, Xidian University, Xi’an, China. (email: chaoweifang@outlook.com)}, \\Dingwen Zhang\thanks{Dingwen Zhang is with the Brain and Artificial Intelligence Laboratory, School of Automation, Northwestern Polytechnical University, Xi’an, China. (email: zhangdingwen2006yyy@gmail.com)},~\IEEEmembership{Member,~IEEE,} 
Meng Wang\textsuperscript{\Envelope}\thanks{Meng Wang is with the Key Laboratory of Knowledge Engineering with Big Data, Ministry of Education, and School of Computer Science and Information Engineering, Hefei University of Technology (HFUT), Hefei,
230601, China, and also with the Hefei Comprehensive National Science Center, Institute of Artificial Intelligence, Hefei, 230026, China. (email: eric.mengwang@gmail.com)}, ~\IEEEmembership{Fellow,~IEEE,}

}

\markboth{IEEE TRANSACTIONS ON MULTIMEDIA}%
{How to Use the IEEEtran \LaTeX \ Templates}

\maketitle

\begin{abstract}
Weakly supervised semantic segmentation (WSSS) has recently attracted considerable attention because it requires fewer annotations than fully supervised approaches, making it especially promising for large-scale image segmentation tasks. Although many vision transformer–based methods leverage self-attention affinity matrices to refine Class Activation Maps (CAMs), they often treat each layer’s affinity equally and thus introduce considerable background noise at deeper layers, where attention tends to converge excessively on certain tokens (i.e., over-smoothing). We observe that this deep-level attention naturally converges on a subset of tokens, yet unregulated query-key affinity can generate unpredictable activation patterns (undisciplined over-smoothing), adversely affecting CAM accuracy. To address these limitations, we propose an Adaptive Re-Activation Mechanism (AReAM), which exploits shallow-level affinity to guide deeper-layer convergence in an entropy-aware manner, thereby suppressing background noise and re-activating crucial semantic regions in the CAMs. Experiments on two commonly used datasets demonstrate that AReAM substantially improves segmentation performance compared with existing WSSS methods, reducing noise while sharpening focus on relevant semantic regions. Overall, this work underscores the importance of controlling deep-level attention to mitigate undisciplined over-smoothing, introduces an entropy-aware mechanism that harmonizes shallow and deep-level affinities, and provides a refined approach to enhance transformer-based WSSS accuracy by re-activating CAMs.
\end{abstract}


\section{Introduction}
\IEEEPARstart{S}{emantic} segmentation is a fundamental task in computer vision that aims to assign a semantic category to each pixel in an image. Although neural networks have achieved remarkable performance on this task, they typically require costly pixel-level annotations. To alleviate this bottleneck, weakly supervised semantic segmentation (WSSS) has been proposed to leverage more easily acquired annotations, such as bounding boxes~\cite{lee2021bbam}, scribbles~\cite{su2022sasformer}, points~\cite{bearman2016point}, or image-level class labels~\cite{2018PSA,2019IRN}. Among these, class labels are particularly accessible as they are readily available in many large-scale datasets. However, because such labels merely indicate the presence or absence of object categories, WSSS with class labels remains challenging. In this work, we focus on WSSS driven by class labels.

Previous methods using class labels generally rely on Class Activation Maps (CAMs)\cite{2016CAM} to localize objects in an image. The resulting pseudo labels serve as ground-truth annotations for fully supervised training, making the quality of both CAMs and pseudo labels critical. CAMs can be generated by either Convolutional Neural Networks (CNN) or Vision Transformers (ViT). Although CNN-based CAMs tend to focus on only the most discriminative regions due to the localized nature of convolution\cite{2021AdvCAM,2022AMR,chen2023extracting}, vision transformers can capture long-range dependencies via self-attention~\cite{dosovitskiy2020ViT}. Consequently, recent approaches~\cite{2022MCTformer,2022AFA,li2023transcam} have used self-attention affinity to refine initial CAMs, achieving significant gains. ToCo~\cite{ru2023token} addresses the \emph{over-smoothing} phenomenon—where token embeddings converge to similar representations after multiple self-attention layers—but we find that over-smoothing itself is not necessarily detrimental to WSSS.

More specifically, our investigation into the query-key affinities of self-attention reveals distinctive patterns across transformer layers: the shallow layers show strong self-correlation (\textit{diagonal} affinity), while deeper layers converge on a narrow subset of tokens (\textit{strip} affinity). Crucially, many of these selected tokens are object-irrelevant, introducing substantial background noise when deep-level affinities are used for CAM refinement. We term this phenomenon \emph{undisciplined over-smoothing}. As discussed in Sec.\ref{sec:analysis}, we further analyze undisciplined over-smoothing both qualitatively and quantitatively, confirming that tokens become more uniform at deeper layers—aligning with existing findings\cite{zhou2021deepvit,zhou2021refiner,touvron2021going,gong2021vision,chen2022principle}. We hypothesize that such convergence is a natural outcome of model training, where vision transformers strategically attend to a limited set of tokens for decision-making. The key issue lies in \emph{where} deep-level attention converges: non-object tokens can dominate, producing excessive background noise in the refined CAMs.

To mitigate this challenge, we propose an \emph{Adaptive Re-Activation Mechanism} (AReAM) that constrains deep-level affinities with shallow-level guidance (Sec.\ref{sec:AReAM}). Unlike recent works\cite{li2023transcam,2022MCTformer} that treat all layers’ affinities equally, AReAM employs entropy-aware \emph{re-activation weights} to dynamically adjust each layer’s contribution to CAM refinement. Shallow-level affinities thus provide primary supervision, while \emph{counterpart weights}—negatively correlated to the re-activation weights—help deep-level affinities converge toward genuine object regions. This simple yet effective design counters undisciplined over-smoothing and markedly enhances WSSS performance.

\textbf{In summary, our contributions are threefold:} 

\begin{itemize} 
\item We conduct an in-depth analysis of how query-key affinities across transformer layers affect CAM refinement, highlighting that undisciplined over-smoothing substantially increases background noise and degrades WSSS results.
\item We introduce AReAM, an adaptive re-activation strategy that leverages shallow-layer affinities to guide deeper-layer attention convergence in an entropy-aware manner, effectively re-activating semantic regions in CAMs. 
\item We demonstrate through extensive experiments on PASCAL VOC 2012 and MS COCO 2014 that our solution not only mitigates undisciplined over-smoothing but also achieves superior results in WSSS tasks.
\end{itemize}
\section{Related Work}
\label{sec:related_works}

\subsection{Weakly Supervised Semantic Segmentation (WSSS)}
Mainstream methods for WSSS with class labels typically employ Class Activation Maps (CAMs) to generate object localization maps that serve as pseudo labels for training fully supervised semantic segmentation networks; however, these often fall short of capturing entire objects. Numerous strategies have been proposed to improve CAM quality, including seed region expansion~\cite{kolesnikov2016seed, huang2018grow}, adversarial erasing~\cite{wei2017erasing, hou2018self_erasing_network, zhang2018adversarial}, online attention accumulation~\cite{jiang2019accumulation, jiang2021accumulation}, cross-image semantic relations~\cite{sun2020mining, li2021group, zhou2022regional, chen2024knowledge, qin2024enhanced, su2024cross}, and contrastive learning between pixels and prototypes~\cite{du2022prototype, chen2022prototype, 10599238,Wu_2024_WACV}. Another line of research refines CAMs based on pairwise affinity within vision transformers~\cite{gao2021TSCAM, sun2021GETAM, 2022AFA, 2022MCTformer, su2022re, li2023transcam, ru2023token, xu2023mctformer+, HAN2025110922, wu2024dupl}. For example, MCTformer~\cite{2022MCTformer, xu2023mctformer+} introduces multiple class tokens to disentangle contextual information among categories, and TransCAM~\cite{li2023transcam} explicitly exploits self-attention weights for semantic propagation. AFA~\cite{2022AFA} supervises inherent self-attention via reliable affinity labels, while ToCo~\cite{ru2023token} aligns patch-token similarities to enhance segmentation accuracy. Distinct from methods such as AFA~\cite{2022AFA} and ToCo~\cite{ru2023token}, we scrutinize query-key affinities in each transformer layer and propose an entropy-aware adaptive re-activation strategy to direct deep-level attention toward semantic regions. WeakTr~\cite{zhu2023weaktr} also highlights how different heads can focus on irrelevant information, introducing a small neural network for weighting—although in a less interpretable way—whereas we delve deeper into refining CAMs to capture object regions more comprehensively. Unlike the above methods, SSC~\cite{10418852} introduces spatial structure constraints that effectively constrain the activation within the object area to alleviate object over-activation and facilitate the activation of more integral object regions, thereby mitigating object under-activation.

\subsection{Over-Smoothing in Vision Transformers}
Despite their remarkable performance across a variety of computer vision tasks, vision transformers have been shown to experience a phenomenon called \emph{over-smoothing} at deeper layers, wherein token representations become nearly identical~\cite{zhou2021deepvit, zhou2021refiner, touvron2021going, gong2021vision}. Chen \textit{et al.}~\cite{chen2022principle} provide a comprehensive examination of this issue by investigating the similarities in embeddings, attention mechanisms, and weights. Building on these insights, we focus on how query-key affinities evolve throughout the layers of a vision transformer and observe that, at deeper layers, each query token consistently attends to a similar subset of key tokens, indicating a convergence toward restricted token sets. According to Darcet \textit{et al.}~\cite{darcet2023vision}, a sufficiently trained vision transformer often detects and leverages \emph{low-informative} tokens to maintain and retrieve global information. Consistent with these findings, our qualitative analyses suggest that deep-layer affinities primarily highlight background regions, thereby introducing noise into tasks such as semantic segmentation. Motivated by this observation, we propose a mechanism to guide deep-layer affinities away from irrelevant tokens and toward object-focused regions, thus mitigating the negative consequences of undisciplined over-smoothing.
\section{Preliminaries}
\label{sec:preliminary}

\noindent In this section, we briefly introduce the method of CAM generation, and how to refine CAMs with the extracted affinities from self-attention. 

\subsection{CAM Generation}
\label{sec:cam_generation}
We employ Class Activation Maps (CAMs)~\cite{2016CAM} to approximately localize objects in an image. Let \(\mathbf{F} \in \mathbb{R}^{D \times H \times W}\) be the feature map derived from the vision transformer, where \(D\) is the embedding dimension and \(H, W\) are spatial dimensions. The original CAM, denoted as \(\mathbf{M} \in \mathbb{R}^{C \times H \times W}\) (with \(C\) the number of object categories), is computed as:
\begin{equation}
\mathbf{M}_{c,:,:} 
= \operatorname{ReLU}\Bigl(\sum_{d=1}^{D} \mathbf{W}_{d,c} \,\mathbf{F}_{d,:,:}\Bigr).
\end{equation}
Here, \(\mathbf{W} \in \mathbb{R}^{D \times C}\) represents the classifier weights, and \(\operatorname{ReLU}(\cdot)\) filters out negative activations. The resulting CAM is then normalized to the range \([0, 1]\) by min-max scaling.

Although CAMs highlight the most discriminative regions, they often fail to capture complete objects—an issue that limits their effectiveness for weakly supervised semantic segmentation. In recent years, transformer-based approaches have shown great promise for modeling global context, which can complement these original CAMs. In particular, a growing number of methods leverage the self-attention affinity matrices from vision transformers to refine CAMs, as will be discussed in the next subsection.

\subsection{CAM Refinement with Affinity Matrices}
\label{sec:cam_refinement}

\begin{figure*}[!htb]
    \centering
    \subfloat[]{\includegraphics[width=0.24\textwidth]{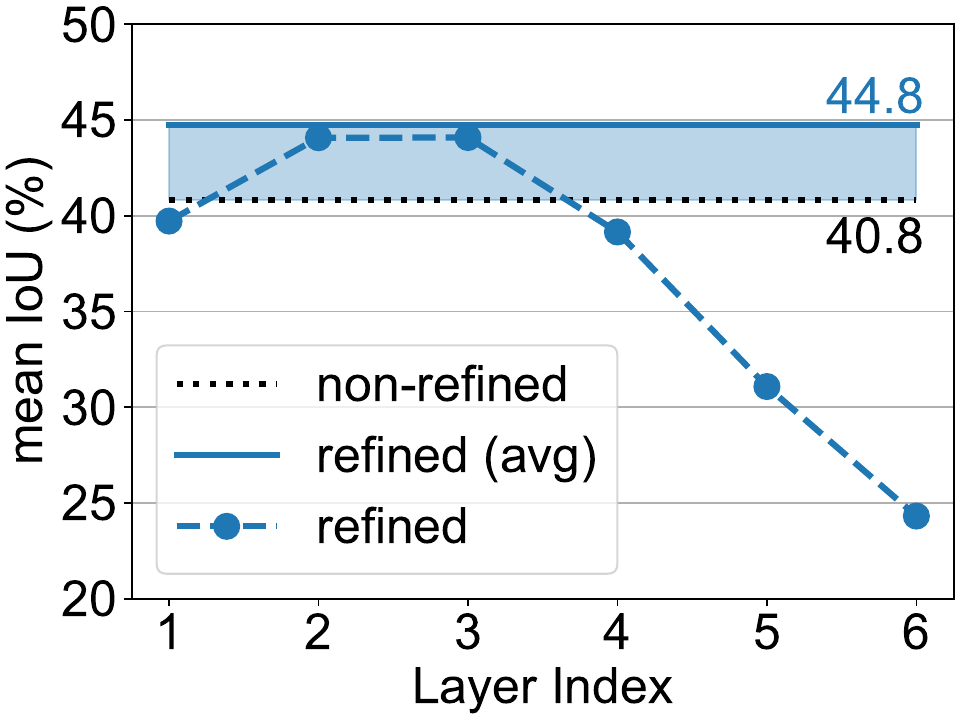}%
    \label{fig:issue_depth6}}
    \hfil
    \subfloat[]{\includegraphics[width=0.24\textwidth]{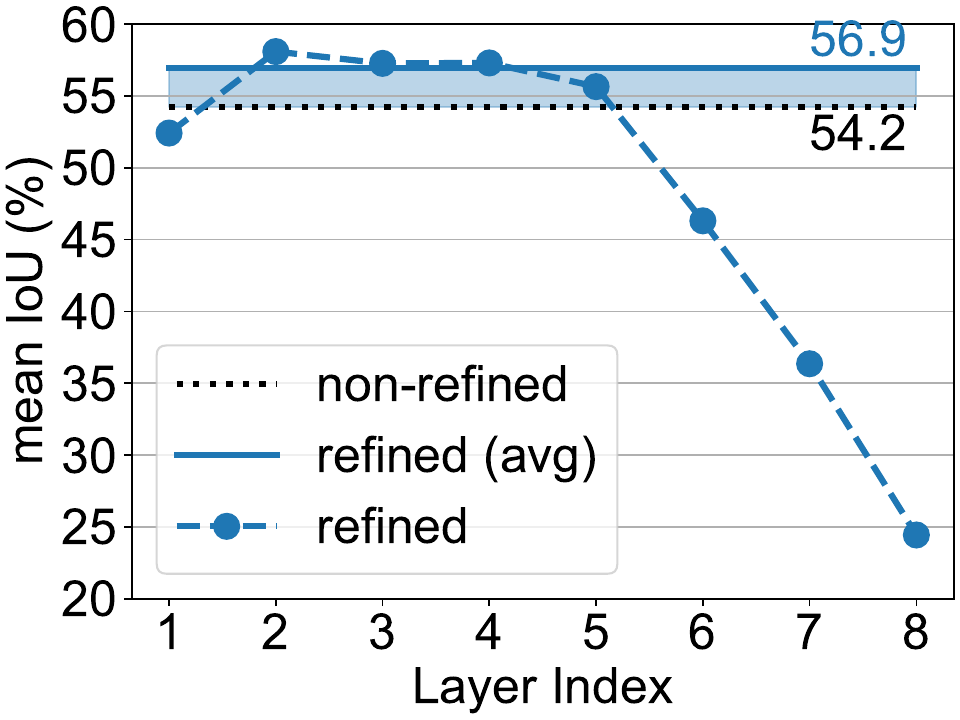}%
    \label{fig:issue_depth8}}
    \hfil
    \subfloat[]{\includegraphics[width=0.24\textwidth]{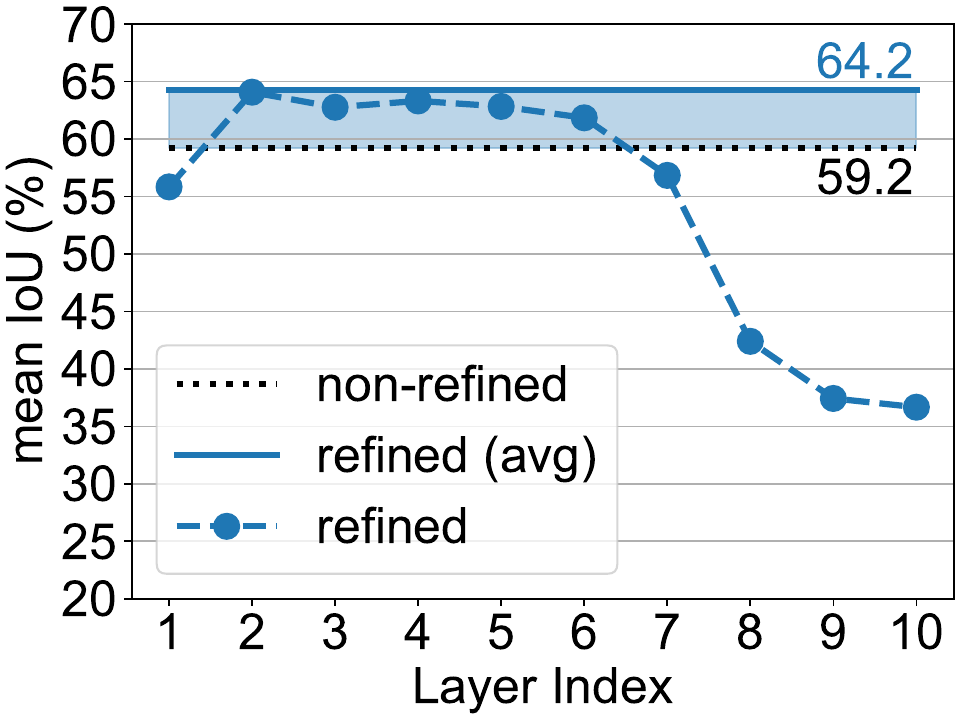}%
    \label{fig:issue_depth10}}
    \hfil
    \subfloat[]{\includegraphics[width=0.24\textwidth]{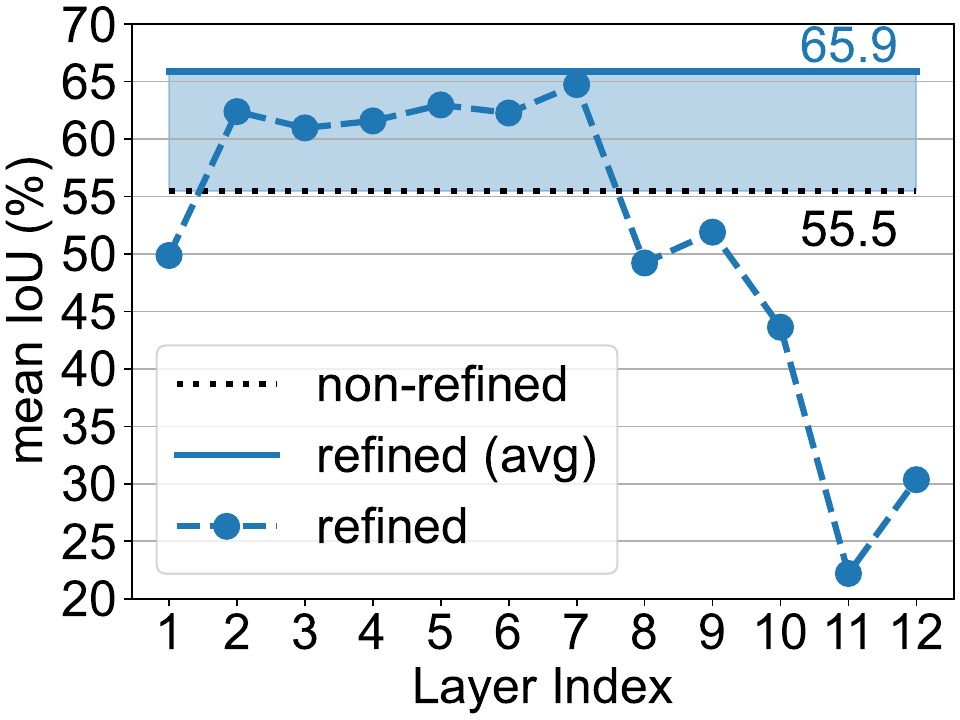}%
    \label{fig:issue_depth12}}
    \caption{We illustrate how affinity matrices at different transformer layers influence CAM refinement by reporting the mIoU (\%) results for a vision transformer with 6, 8, 10, or 12 total layers in each sub-figure. In each scenario, the dotted horizontal line (\emph{non-refined}) denotes the original CAM, while the solid horizontal line (\emph{refined (avg)}) represents the refined CAM $\bar{\mathbf{M}}$, obtained by aggregating the average affinity matrix $\bar{\mathbf{A}}$ from all layers. The blue shaded area highlights the performance gain over the non-refined baseline, and the dashed lines (\emph{refined}) provide layer-wise insights, showing that shallow-layer affinity notably improves CAM quality whereas deeper-layer affinity can diminish it. Overall, these results underscore the importance of query-key affinities in enhancing CAMs and reveal that effectively leveraging global relationships across layers remains an open challenge.}
    \label{fig:teaser}
\end{figure*}

\begin{figure}[!htb]
    \centering
    \includegraphics[width=1.0\linewidth]{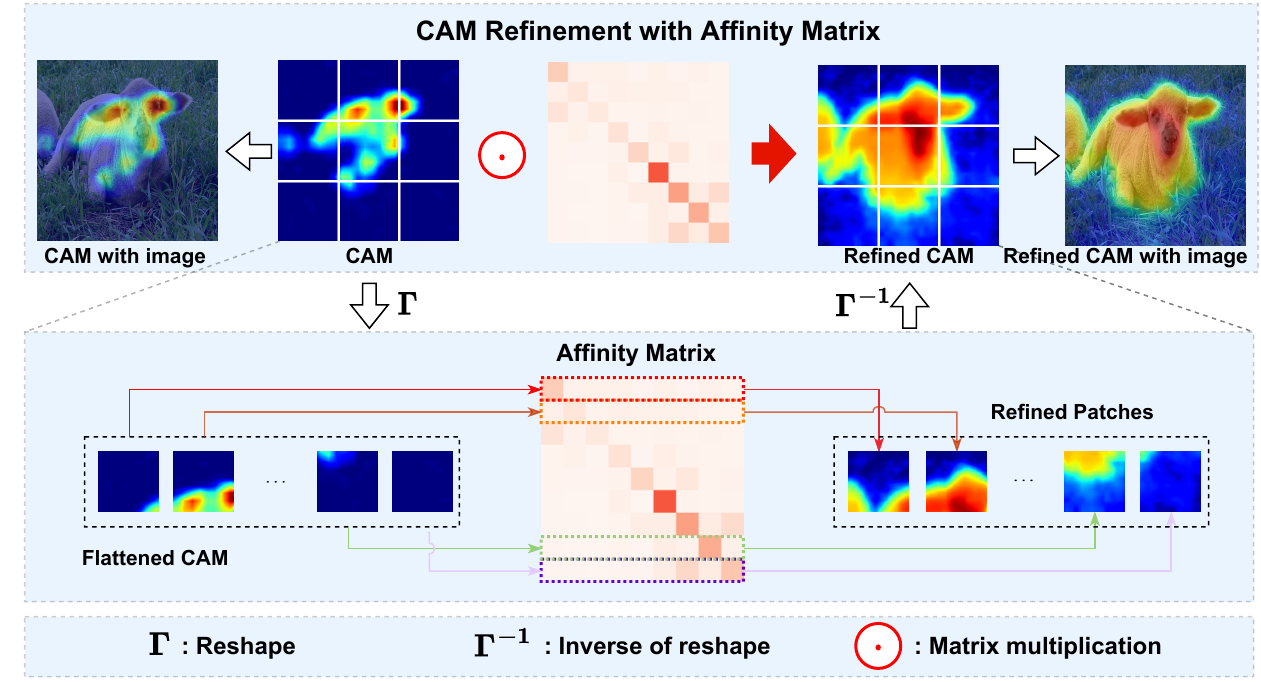}
    \caption{Affinity-based CAM refinement.}
    \label{fig:propagation}
\end{figure}

An input image is first subdivided into $H \times W$ patches, which are flattened and linearly projected into a sequence of $N = H \times W$ tokens. Transformers use the self-attention mechanism~\cite{vaswani2017attention} to model long-range dependencies among these tokens. Specifically, at each transformer layer $l$, the token sequence is mapped into query, key, and value matrices 
\(\mathbf{Q}^{(l)}, \mathbf{K}^{(l)}, \mathbf{V}^{(l)} \in \mathbb{R}^{N \times D}\). 
The scaled dot-product attention~\cite{vaswani2017attention} computes the \emph{affinity matrix} \(\mathbf{A}^{(l)} \in \mathbb{R}^{N \times N}\) by taking dot products of the queries and keys, dividing by \(\sqrt{D}\), and applying a \(\mathtt{softmax}\). The final output \(\mathbf{O}^{(l)} \in \mathbb{R}^{N \times D}\) is then a weighted sum of the values, where the weights are given by the softmax-normalized affinities (see Fig.~\ref{fig:propagation}):
\begin{align}
    \mathbf{A}^{(l)} &= \frac{\mathbf{Q}^{(l)} \bigl(\mathbf{K}^{(l)}\bigr)^\top}{\sqrt{D}}, \\
    \mathbf{O}^{(l)} &= \sigma\bigl(\mathbf{A}^{(l)}\bigr) \, \mathbf{V}^{(l)},
\end{align}
where \(\sigma(\cdot)\) is the \(\mathtt{softmax}\) function. To fuse the affinity information across different layers, previous methods simply average the softmax-normalized affinity matrices. Formally, the \emph{average affinity matrix} \(\bar{\mathbf{A}} \in \mathbb{R}^{N \times N}\) is defined as:
\begin{equation}
    \bar{\mathbf{A}} = \frac{1}{L} \sum_{l=1}^{L} \sigma\bigl(\mathbf{A}^{(l)}\bigr),
\end{equation}
where \(L\) is the total number of transformer layers. The refined CAM \(\bar{\mathbf{M}} \in \mathbb{R}^{C \times H \times W}\) is then obtained by reshaping each 2D map \(\mathbf{M}_{c,:,:}\) into a 1D sequence, multiplying by \(\bar{\mathbf{A}}^\top\), and reshaping back:
\begin{equation}
    \bar{\mathbf{M}}_{c,:,:} = \Gamma^{-1} \Bigl(\Gamma \bigl(\mathbf{M}_{c,:,:}\bigr) \cdot \bar{\mathbf{A}}^{\top}\Bigr),
\end{equation}
where \(\Gamma(\cdot)\) flattens a 2D map into a 1D sequence, and \(\Gamma^{-1}(\cdot)\) restores the 2D spatial dimensions.

\noindent \textbf{Remark.} Fig.~\ref{fig:teaser} illustrates how CAM refinement through transformer affinity matrices affects performance. We consider four scenarios where the total number of transformer layers differs. In each sub-figure, the dotted horizontal line (\emph{non-refined}) represents the original CAM, while the solid horizontal line (\emph{refined (avg)}) shows the result of the refined CAM $\bar{\mathbf{M}}$, obtained by aggregating the average affinity matrix $\bar{\mathbf{A}}$ across all layers. The blue shaded area highlights the performance gain of the refined CAM. We further provide layer-wise refinements (marked as \emph{refined}) using blue dashed lines and markers, demonstrating the contribution of each layer’s affinity to CAM enhancement. Notably, while shallow-layer affinity substantially improves the original CAM, deeper-layer affinity can degrade its performance. In summary, the query-key affinity is instrumental for refining CAMs, though its efficacy is uneven across layers. Understanding and harnessing these global relationships remains an open challenge, meriting further exploration.

\section{Method}
\label{sec:method}
\noindent Building on the insights outlined above, we propose an adaptive solution in Sec.\ref{sec:AReAM}, guided by the in-depth analysis of affinities presented in Sec.\ref{sec:analysis}.

\subsection{Analysis of Affinity Matrices}
\label{sec:analysis}

\begin{figure*}
    \centering
    \resizebox{0.95\linewidth}{!}{
        \includegraphics[height=3cm]{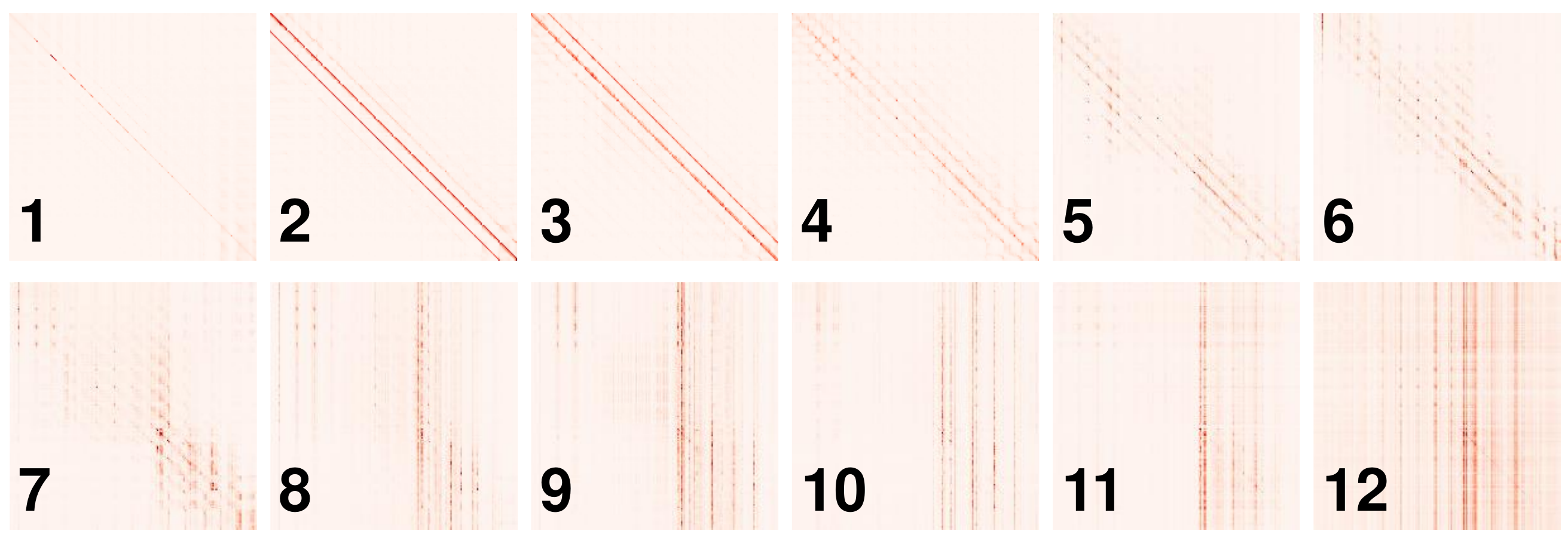}
        \quad
        \includegraphics[height=3cm]{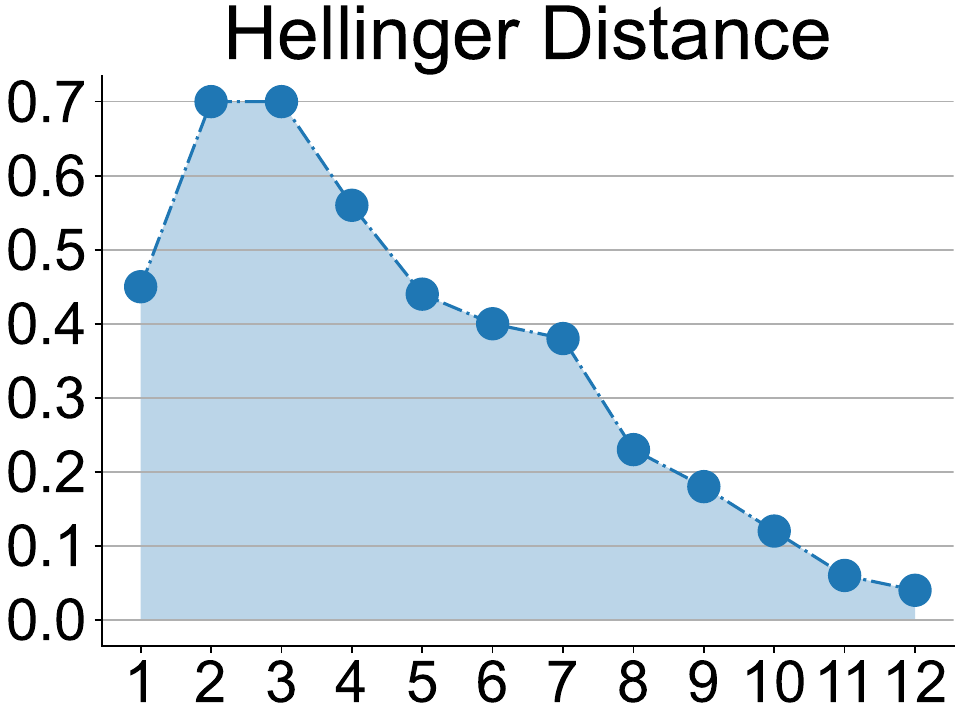}
    }
    \caption{ Analysis of the affinity matrix. Each row represents a query token while each column a key token. (left) Visualization of the normalized affinity matrix in each layer. (right) Hellinger distance within the normalized affinity matrix in each layer. }
    \label{fig:affinity_matrix_visualization}
\end{figure*}

\noindent \textbf{Attention Convergence}.
While the global relationships captured by a single-layer affinity do not always enhance CAMs, we explore the underlying cause by examining the affinity matrices layer by layer. As illustrated in Fig.~\ref{fig:affinity_matrix_visualization} (left), shallow layers (e.g., layers 1--6) predominantly exhibit diagonal correlations, indicating strong self-similarity. In contrast, deeper layers (e.g., layers 11--12) converge on a restricted subset of key tokens, reflecting a narrower attention focus. This shift in attention patterns helps explain the inconsistent efficacy of layer-wise affinities for CAM refinement.

To further investigate how the affinity matrix affects the WSSS task, we measure the distributional differences in affinity across layers using the Hellinger distance. Let 
\(\tilde{\mathbf{A}}^{(l)} = \sigma\bigl(\mathbf{A}^{(l)}\bigr) \in \mathbb{R}^{N \times N}\) 
denote the normalized affinity matrix at layer \(l\). The Hellinger distance between the \(i\)-th and \(j\)-th distributions is defined as:
\begin{equation}
    \mathcal{H}_l\bigl(\tilde{\mathbf{A}}^{(l)}_{i,:}, \tilde{\mathbf{A}}^{(l)}_{j,:}\bigr)
    = \frac{1}{\sqrt{2}} \sqrt{\sum_{k=1}^{N} \Bigl(\sqrt{\tilde{\mathbf{A}}^{(l)}_{i,k}} 
    - \sqrt{\tilde{\mathbf{A}}^{(l)}_{j,k}}\Bigr)^2}.
\end{equation}
We compute this distance using a 12-layer vision transformer fine-tuned on the PASCAL VOC 2012 dataset. As shown in Fig.~\ref{fig:affinity_matrix_visualization} (right), the distances steadily decrease as the layer index increases, indicating that attention converges on a narrower set of key tokens at deeper layers.

\begin{figure*}
    \centering
    \includegraphics[width=0.95\textwidth]{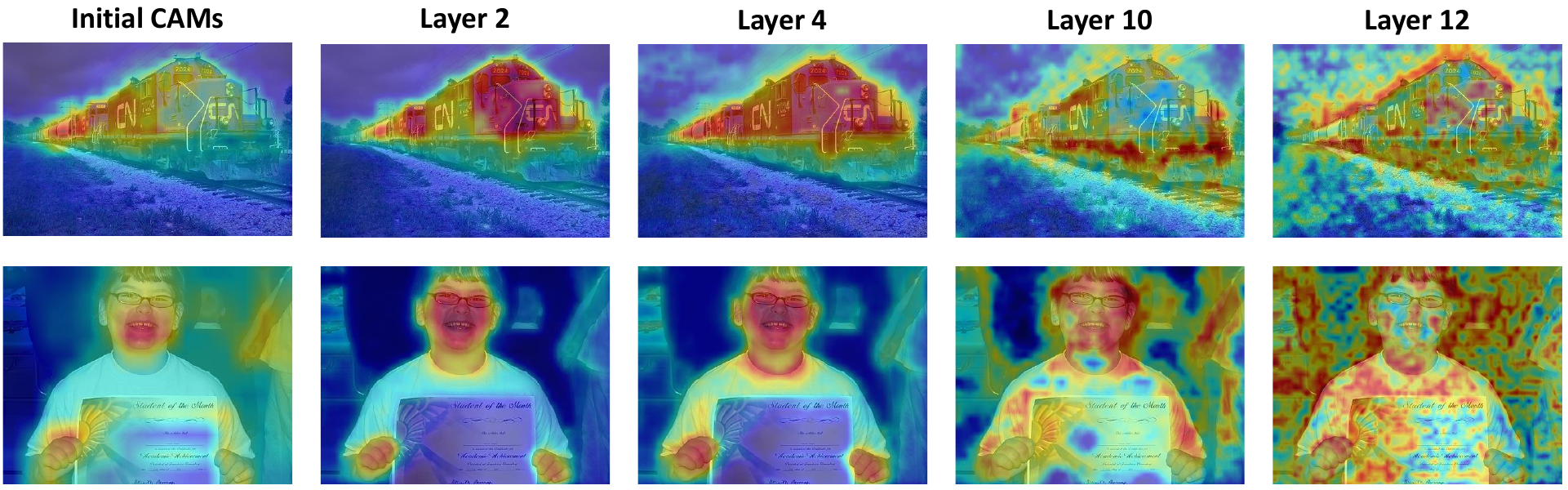}
    \caption{\centering Examples of refined CAMs utilizing the affinity matrix at shallower layers (2, 4) and at deeper layers (10, 12).}
    \label{fig:bg_noise}
\end{figure*}

\noindent \textbf{Undisciplined Over-Smoothing Brings Global Noise.}
Our findings reveal that, at sufficiently deep transformer layers, each query token persistently attends to a similar subset of key tokens (i.e., over-smoothing). To examine how this affects the original CAM, we refer to Fig.~\ref{fig:bg_noise}, which shows that the transition from shallower to deeper layers progressively contaminates the refined CAM with undesired global noise.

We posit that over-smoothing in query-key affinities is a \emph{normal} outcome of well-trained vision transformers. More importantly, we attribute the global noise to \emph{undisciplined} over-smoothing, wherein deep-level attention converges on key tokens that predominantly represent background rather than foreground objects. Empirical validation of this assumption is presented in Section~\ref{sec:experiment}.

In summary, both qualitative and quantitative evidence suggests that as the depth increases, each query token converges on an increasingly narrow subset of key tokens, indicating deeper-layer attention convergence. Simultaneously, these undisciplined affinities inject background noise into the refined CAMs.

\noindent \textbf{Remark.}
Although earlier studies~\cite{li2023transcam, 2022MCTformer} utilize transformer affinities to refine CAMs, they simply average the affinity matrices across all layers, overlooking the potential over-smoothing at deeper layers. While some works~\cite{gong2021vision, chen2022principle, ru2023token} attribute performance degradation directly to over-smoothing in vision transformers, we argue that over-smoothing itself is a natural characteristic of deep-level attention. Instead, we identify \emph{undisciplined over-smoothing}—where attention erroneously emphasizes background tokens—as the primary source of non-semantic noise in refined CAMs, ultimately causing substantial declines in WSSS performance.

\begin{figure*}[!htb]
\centering
\includegraphics[width=0.95\textwidth]{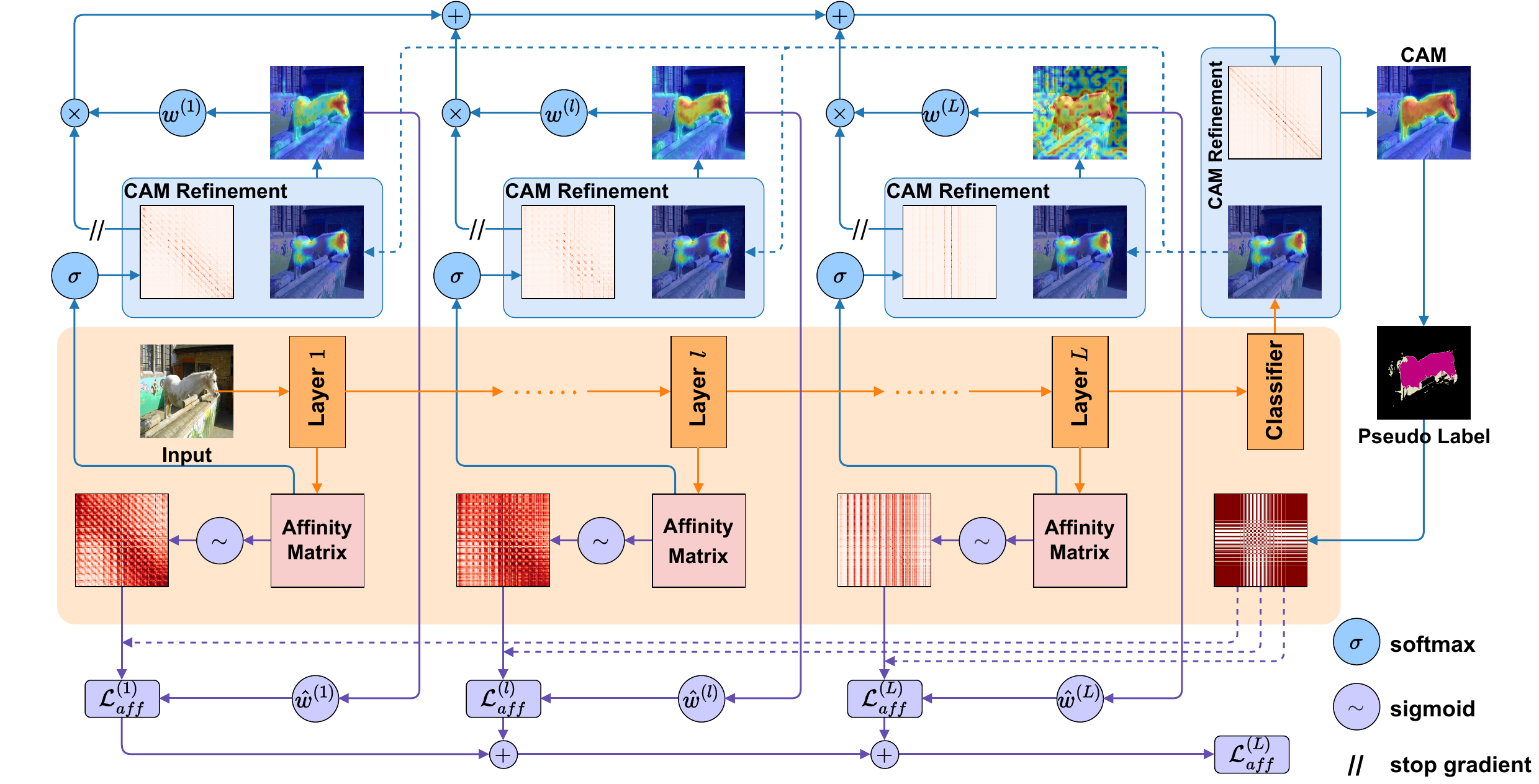}
    \caption{The overview of the proposed AReAM. At the training stage, AReAM applies entropy-aware $w^{(l)}$ to the affinity matrix of each layer to adjust its contribution. To address undisciplined over-smoothing, we supervise the affinity matrix of each layer with $\hat{w}^{(l)}$ to instruct attended tokens to converge to semantic objects. At the inference stage, the disciplined affinity matrices of all layers are averaged to refine CAMs.}
    \label{fig:overview}
\end{figure*}

\subsection{Adaptive Re-Activation Mechanism}
\label{sec:AReAM}

\noindent \textbf{Overview.}
Fig.~\ref{fig:overview} illustrates the overall architecture of our proposed AReAM. We adopt a vision transformer as the backbone, supervised primarily by image-level class labels. Following the method in Section~\ref{sec:cam_generation}, we first generate CAMs from the extracted feature maps to roughly localize objects of interest. We then refine these initial CAMs with the affinity matrices via a random-walk procedure within AReAM to capture more complete object regions.

To accommodate a vision transformer with $L$ layers, we introduce \emph{re-activation weights} $\{w^{(l)} \mid l=1,\dots,L\}$ to regulate each layer’s affinity contribution to CAM refinement. These weights are computed based on the \emph{entropy} of the refined CAM, where higher entropy indicates stronger undisciplined over-smoothing. In other words, greater randomness (i.e., higher entropy) implies that a layer’s affinity matrix is less beneficial, yielding a lower re-activation weight.

In tandem, we define a set of \emph{counterpart weights} $\{\hat{w}^{(l)} \mid l=1,\dots,L\}$, negatively correlated with the re-activation weights, to facilitate affinity optimization. By assigning larger counterpart weights to layers with higher entropy, we encourage these layers to focus attention on more semantically meaningful tokens. Concretely, re-activation weights $w^{(l)}$ aggregate affinity relations for refining CAMs, while counterpart weights $\hat{w}^{(l)}$ aggregate affinity losses, with each pair satisfying $w^{(l)} + \hat{w}^{(l)} = 1$ (Eq.~(10)). If a layer’s refined CAM exhibits pronounced noise ($e^{(l)} \uparrow$), its contribution to CAM refinement is reduced ($w^{(l)} \downarrow$), whereas its counterpart weight ($\hat{w}^{(l)} \uparrow$) increases, encouraging the network to redirect attention away from background tokens. As training proceeds, both sets of weights adapt dynamically to diminish global noise introduced by undisciplined over-smoothing (see Fig.~\ref{fig:layerwise_visualization} for visual examples).

At inference time, we simply average the \emph{disciplined} affinity matrices across all layers to finalize token-wise relations for CAM refinement.

\noindent \textbf{CAM Refinement within AReAM.}
Recall from Section~\ref{sec:preliminary} that \(\mathbf{M} \in \mathbb{R}^{C \times H \times W}\) denotes the original CAM, and each layer’s affinity matrix is represented by \(\mathbf{A}^{(l)} \in \mathbb{R}^{HW \times HW}\). To derive the two sets of weights, we first generate a refined CAM \(\mathbf{M}^{(l)}\) for each layer by combining the original CAM with its corresponding affinity matrix, and then compute the entropy \(e^{(l)} \in \mathbb{R}\) of the refined CAM:
\begin{align}
    \mathbf{M}^{(l)}_{c,:,:} &= \Gamma^{-1}\bigl(\Gamma(\mathbf{M}_{c,:,:}) \cdot \sigma(\mathbf{A}^{(l)})^\top\bigr), \\
    e^{(l)} &= -\sum_{h,w} \sigma\Bigl(\max_c \mathbf{M}^{(l)}_{c,h,w}\Bigr) \,\log \sigma\Bigl(\max_c \mathbf{M}^{(l)}_{c,h,w}\Bigr).
\end{align}

Here, we convert each \(\mathbf{M}^{(l)}\) into a class-agnostic object heatmap by taking the maximum along the category dimension, acknowledging that an image may contain multiple classes. Next, we define the re-activation weights \(w^{(l)}\) and counterpart weights \(\hat{w}^{(l)}\) as follows:
\begin{align}
    \hat{w}^{(l)} &= \frac{e^{(l)} - \min_l e^{(l)}}{\max_l e^{(l)} - \min_l e^{(l)}}, \\
    w^{(l)} &= 1 - \hat{w}^{(l)}.
\end{align}

The re-activation weights are then used to combine the layer-wise affinity matrices into a single matrix \(\hat{\mathbf{A}}\), which is ultimately applied to the original CAM to produce the re-activated CAM \(\hat{\mathbf{M}}\):
\begin{align}
    \hat{\mathbf{A}} &= \sum_{l=1}^{L} w^{(l)} \,\sigma\bigl(\mathbf{A}^{(l)}\bigr), \\
    \hat{\mathbf{M}}_{c,:,:} &= \Gamma^{-1}\Bigl(\Gamma(\mathbf{M}_{c,:,:}) \cdot \hat{\mathbf{A}}^{\top}\Bigr). \label{eq:mat_mul}
\end{align}

Finally, the matrix multiplication in Eq.~(\ref{eq:mat_mul}) serves to re-activate each token by incorporating contextual semantics. Specifically, each row of \(\hat{\mathbf{A}}\) captures the attention distribution of a single token over all other tokens (including itself), and this distribution is used to adaptively enhance the original CAM.

\noindent \textbf{Generate Reliable Pseudo Labels.}
To further enhance pixel-level contextual consistency, existing approaches~\cite{2022MCTformer,li2023transcam} typically adopt dense CRF~\cite{krahenbuhl2011efficient} to incorporate low-level image features. Other frameworks~\cite{araslanov2020single, 2022AFA} employ pixel-adaptive convolution~\cite{su2019pixel} as a considerably faster alternative to dense CRF for refining segmentation labels during training. Drawing inspiration from the Pixel-Adaptive Refinement module (PAR)~\cite{su2019pixel, 2022AFA}, we generate piecewise-smooth results as follows:
\begin{equation}
    \mathbf{Y}(:,i,j) = \sum_{(x,y) \in \mathcal{N}(i,j)} \kappa(i,j,x,y) \, \hat{\mathbf{M}}(:,x,y),
\end{equation}
where $\mathcal{N}(i,j)$ denotes the neighbor set for pixel $(i,j)$, and $\kappa(i,j,x,y)$ is a kernel capturing pairwise potentials between $(i,j)$ and $(x,y)$. In line with~\cite{2022AFA}, we incorporate image intensities and spatial locations to define these pairwise terms:
\begin{align}
    \begin{split}
        \kappa_I(i,j,x,y) &= -\frac{\bigl|I(i,j)-I(x,y)\bigr|^2}{\bigl(c_I \sigma_I\bigr)^2}, \\
        \kappa_L(i,j,x,y) &= -\frac{\bigl|L(i,j)-L(x,y)\bigr|^2}{\bigl(c_L \sigma_L\bigr)^2},
    \end{split}
\end{align}
where $I(i,j)$ and $L(i,j)$ represent the image intensity and spatial location of pixel $(i,j)$, $\sigma_I$ and $\sigma_L$ are locally computed standard deviations, and $c_I$ and $c_L$ control color similarity and spatial proximity. Intuitively, neighboring pixels with similar intensities are likely to share the same class label. The kernel $\kappa(i,j,x,y)$ is thus formed by combining $\kappa_I$ and $\kappa_L$:
\begin{equation}
    \kappa(i,j,x,y) 
    = w_I \,\frac{\exp(\kappa_I)}{\sum_{(x,y)} \exp(\kappa_I)}
    + w_L \,\frac{\exp(\kappa_L)}{\sum_{(x,y)} \exp(\kappa_L)},
\end{equation}
where $w_I$ and $w_L$ are weight factors that adjust the influence of each term. We extract neighbors for each pixel using multiple dilated $3 \times 3$ convolution kernels, running for 10 iterations to balance training efficiency and segmentation accuracy.

To convert the calibrated CAMs into semantic segmentation pseudo labels, we define two thresholds, $\alpha_1$ and $\alpha_2$, satisfying $0 < \alpha_1 < \alpha_2 < 1$. Following \cite{2022AFA}, we then categorize each pixel as reliable foreground, reliable background, or ignored:
\begin{equation}
\small
    \mathbf{Y}_s(i,j)=\left\{
    \begin{array}{lr}
    \operatorname{argmax}\bigl(\mathbf{Y}(:,i,j)\bigr), & \text{if } \max\bigl(\mathbf{Y}(:,i,j)\bigr) > \alpha_2, \\
    0, & \text{if } \max\bigl(\mathbf{Y}(:,i,j)\bigr) < \alpha_1, \\
    255, & \text{otherwise},
    \end{array}
    \right.
\end{equation}
where $\operatorname{argmax}(\cdot)$ obtains the class label with the highest score, $0$ denotes the background, and $255$ indicates uncertain regions.

\noindent \textbf{AReAM Optimization.}
The reliable segmentation labels $\mathbf{Y}_s \in \mathbb{R}^{H \times W}$ are used to derive pixel-wise affinity labels $\mathbf{Y}_a \in \mathbb{R}^{HW \times HW}$ as follows:
\begin{equation}
\small
    \mathbf{Y}_a(ij,kl)=\left\{
    \begin{array}{lr}
    1, & \text{if } \mathbf{Y}_s(i,j) = \mathbf{Y}_s(k,l) \text{ and } \mathbf{Y}_s(:,:) \neq 255, \\
    0, & \text{if } \mathbf{Y}_s(i,j) \neq \mathbf{Y}_s(k,l) \text{ and } \mathbf{Y}_s(:,:) \neq 255, \\
    255, & \text{otherwise}.
    \end{array}\right.
\end{equation}

Specifically, for two pixels $(i,j)$ and $(k,l)$ that both lie in confidently predicted foreground or background regions, the affinity label is $1$ if they share the same class, and $0$ otherwise. Any pixel pair involving an uncertain region (label $255$) is ignored during optimization. We treat $\mathbf{Y}_a$ as the ground truth for supervising each layer’s affinity matrix $\mathbf{A}^{(l)}$. The corresponding affinity loss $\mathcal{L}_{aff}$, weighted by our counterpart weights $\hat{w}^{(l)}$, is defined by:
\begin{align}
\small
    \mathcal{L}_{aff}^{(l)} &= \frac{1}{N^{+}} \sum_{\mathbf{Y}_{a}(ij,kl)=1} \Bigl( 1 - \tilde{\sigma}\bigl(\mathbf{A}^{(l)}_{ij,kl}\bigr)\Bigr) \nonumber\\
    &\quad + \frac{1}{N^{-}} \sum_{\mathbf{Y}_{a}(ij,kl)=0} \tilde{\sigma}\bigl(\mathbf{A}^{(l)}_{ij,kl}\bigr), \\
    \mathcal{L}_{aff} &= \sum_{l}\hat{w}^{(l)}\,\mathcal{L}_{aff}^{(l)},
    \label{eq:aff_loss}
\end{align}
where $\tilde{\sigma}(\cdot)$ is the $\mathtt{sigmoid}$ function, and $N^{+}$ and $N^{-}$ are the numbers of positive and negative pixel pairs, respectively. Unlike $\mathtt{softmax}$, using $\mathtt{sigmoid}$ here mimics a gating mechanism that encourages consistent affinity for pixel pairs belonging to the same category, thereby facilitating their mutual activation propagation. In this manner, deeper layers exhibiting higher activation entropy receive stronger supervision. Finally, the overall training loss $\mathcal{L}$ combines the classification loss $\mathcal{L}_{cls}$ and the affinity loss $\mathcal{L}_{aff}$ with a weighting coefficient $\lambda$:
\begin{equation}
    \mathcal{L} = \mathcal{L}_{cls} + \lambda\,\mathcal{L}_{aff}.
    \label{eq:total_loss}
\end{equation}

Our proposed AReAM thus serves as a plug-and-play method to mitigate undisciplined over-smoothing in WSSS. Its re-activation and counterpart weights adaptively update during training according to activation randomness, introducing few additional hyperparameters.

\section{Experiments}
\label{sec:experiment}

\noindent In this section, we describe the experimental setup in Section~\ref{sec:setup}, present detailed ablation studies and analyses in Section~\ref{sec:ablation}, and compare our method with state-of-the-art approaches in Section~\ref{sec:SOTA}.

\subsection{Setup}
\label{sec:setup}

\noindent \textbf{Datasets and Evaluation Metrics.}
We evaluate our method on the PASCAL VOC 2012~\cite{pascal} and MS COCO 2014~\cite{lin2014microsoft} datasets. PASCAL VOC 2012 contains 20 object categories with 1,464 \textit{train}, 1,449 \textit{val}, and 1,456 \textit{test} images. Following common practice~\cite{2018PSA,2019IRN,2020SEAM,2021RIB,2022MCTformer}, we use an augmented training set of 10,582 images with image-level annotations from~\cite{SBD}. MS COCO 2014 comprises 80 object categories, including 82,081 \textit{train} and 40,137 \textit{val} images. We only use image-level class labels throughout training, and report mean Intersection-over-Union (mIoU) as the evaluation metric.

\noindent \textbf{Implementation Details.}
We adopt a 12-layer DeiT-S~\cite{2021deit} model pretrained on ImageNet~\cite{russakovsky2015imagenet} as our backbone. Following \cite{2022MCTformer}, we employ multiple class tokens and fuse the class-specific attentions from the last three layers to form the default non-refined CAMs. Data augmentation includes random cropping, random horizontal flipping, and random color jittering. The input size is $448 \times 448$, as in \cite{ru2023token}. We largely follow the training recipe of \cite{2021deit}, with a batch size of 32 and a total of 45 epochs. The classification loss $\mathcal{L}_{cls}$ is the commonly adopted multi-label soft margin loss.

Following established practice~\cite{wang2020equivariant,2021AdvCAM,su2021context,2021CPN,2022MCTformer} in the weakly supervised semantic segmentation (WSSS) community, we apply CRF~\cite{krahenbuhl2011efficient} and PSA~\cite{2018PSA} post-processing on the re-activated CAMs to further improve pseudo-label quality. To generate pseudo labels, we follow the setting in \cite{2022AFA} with $c_I=0.3$, $c_L=0.3$, $w_I=1.0$, $w_L=0.01$, and background thresholds $\alpha_1=0.25$ and $\alpha_2=0.7$.

For semantic segmentation, we experiment with DeepLab-v1 (ResNet38 backbone)~\cite{chen2014semantic} and DeepLab-v2 (ResNet101 backbone)~\cite{deeplab}, following the training protocols of \cite{2022MCTformer}. During inference, we employ multi-scale testing and CRF post-processing~\cite{2022MCTformer,ru2023token}.

\begin{figure*}[!t]
    \centering
    \subfloat[]{\includegraphics[width=0.24\textwidth]{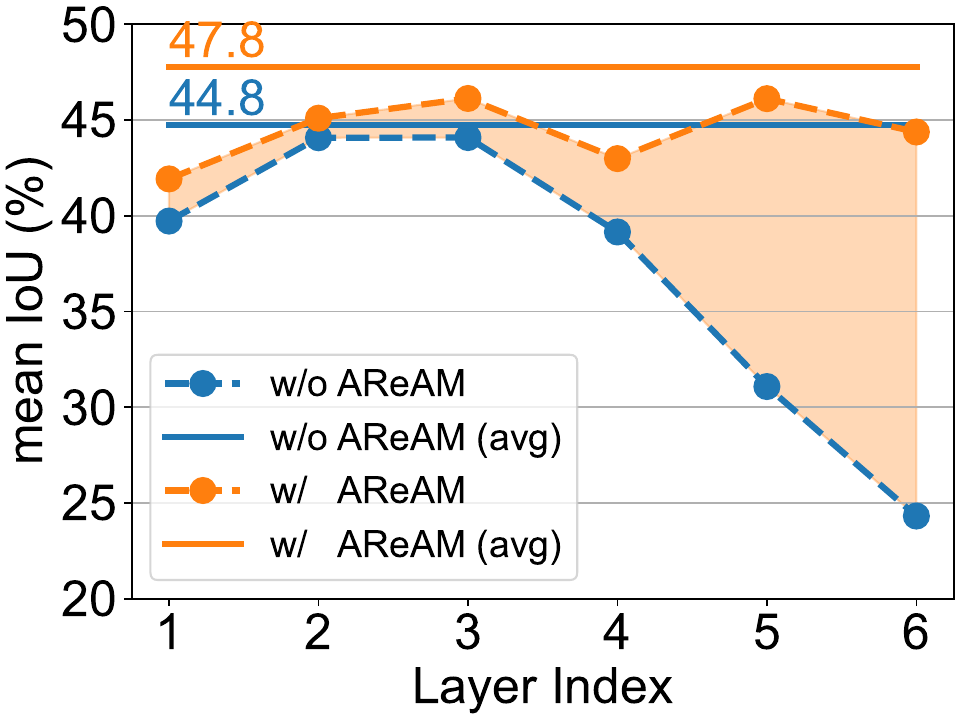}}
    \hfil
    \subfloat[]{\includegraphics[width=0.24\textwidth]{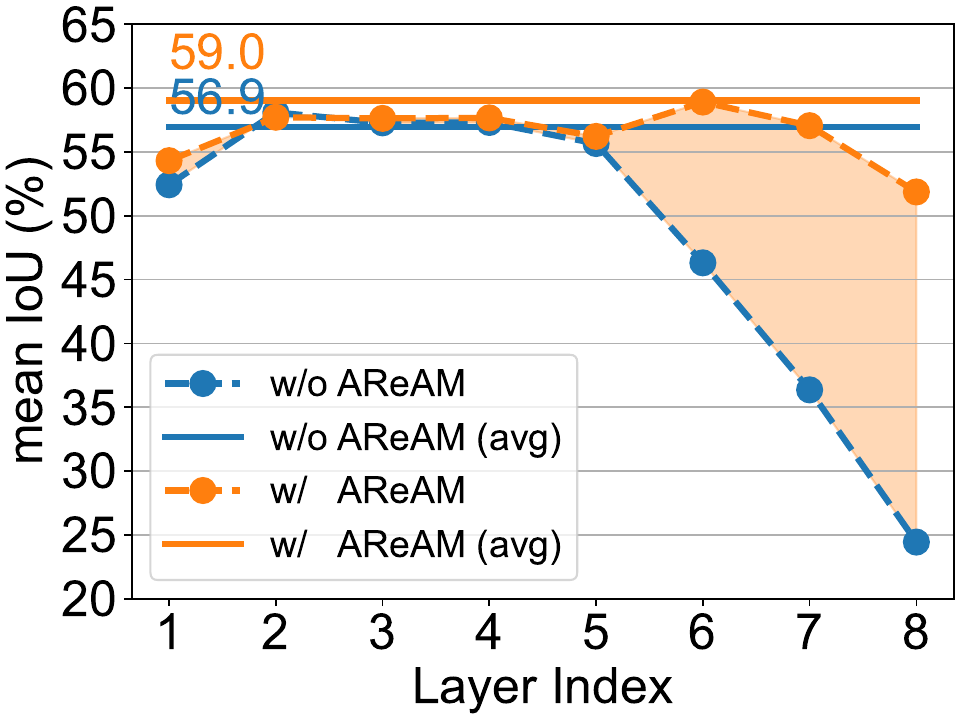}}
    \hfil
    \subfloat[]{\includegraphics[width=0.24\textwidth]{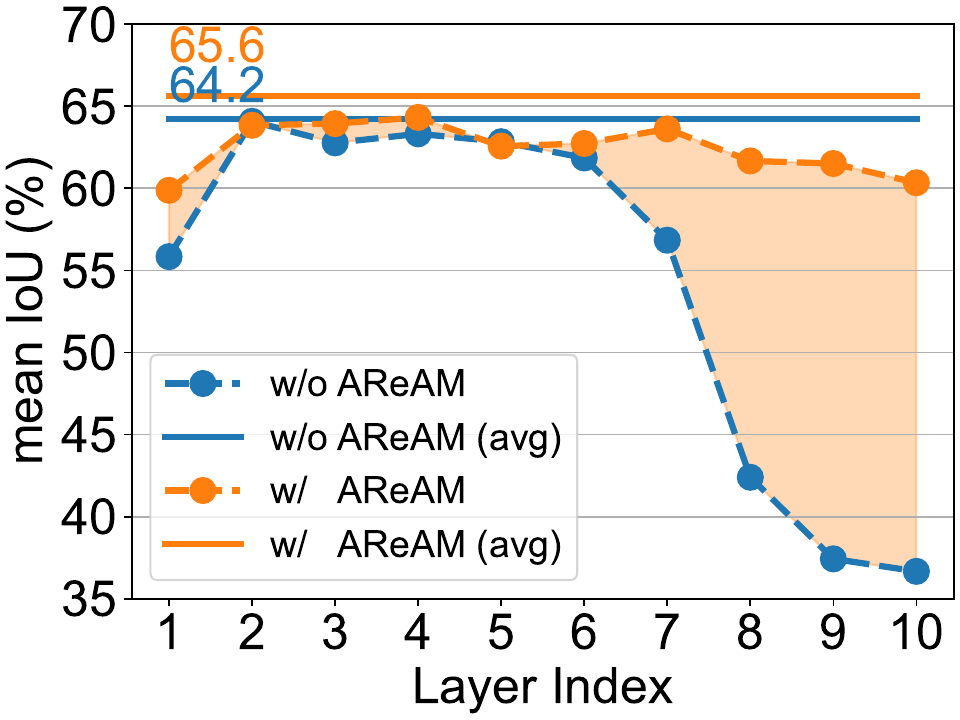}}
    \hfil
    \subfloat[]{\includegraphics[width=0.24\textwidth]{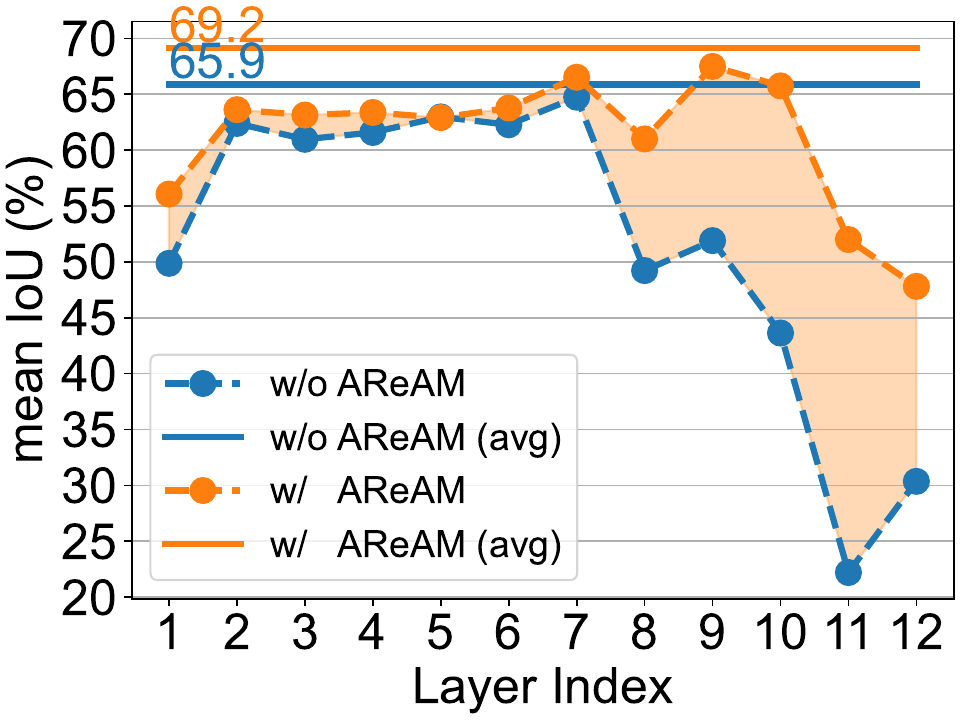}}
    \caption{Layer-wise gains brought by AReAM when the total number of layers are 6, 8, 10, and 12.}
    \label{fig:layerwise_gain}
\end{figure*}

\subsection{Ablation Study and Analysis}
\label{sec:ablation}

\begin{table}[t]
    \centering
    \caption{Performance of pseudo labels on the PASCAL VOC train set in terms of mIoU (\%). ``CAMs'' means non-refined CAMs; ``Affinity'' means refining CAMs with the average affinity matrix; ``P.P.'' means post-processing, a common practice applied to promote pseudo labels.}
    \label{tab:pseudo_label_result}
    \begin{tabular}{l|cc}
        \toprule
        Method & P.P.($\times$) & P.P.(\checkmark) \\
        \midrule
        AdvCAM~\cite{2021AdvCAM}$_\text{CVPR'2021}$         & 55.6 & 68.0 \\
        RIB~\cite{2021RIB}$_{\text{NeurIPS'2021}}$          & 56.5 & 68.6 \\
        CDA~\cite{su2021context}$_{\text{ICCV'2021}}$       & 58.4 & 66.4 \\
        ESOL~\cite{li2022expansion}$_\text{NeurIPS’2022}$   & 53.6 & 68.7 \\
        AMR~\cite{2022AMR}$_{\text{AAAI'2022}}$             & 56.8 & 69.7 \\
        W-OoD~\cite{lee2022weakly}$_\text{CVPR'2022}$       & 59.1 & 72.1 \\
        MCTformer~\cite{2022MCTformer}$_{\text{CVPR'2022}}$ & 61.7 & 69.1 \\
        ViT-PCM~\cite{rossetti2022max}$_{\text{ECCV'2022}}$ & 67.7 & 71.4 \\
        LPCAM~\cite{chen2023extracting}$_\text{CVPR'2023}$  & 54.9 & 71.2 \\
        ACR~\cite{kweon2023weakly}$_\text{CVPR'2023}$       & 65.5 & 70.9 \\
        USAGE~\cite{jo2023mars}$_\text{ICCV'2023}$          & 67.7 & 72.8 \\
        SSC~\cite{10418852}$_\text{IP'2024}$                & 58.3 & 71.9 \\
        COBRA~\cite{HAN2025110922}$_\text{PR'2024}$         & 62.3 & 73.5 \\
        CPMC~\cite{10599238}$_\text{TCSVT'2024}$           & 66.0 & 73.6 \\
        KTSE~\cite{chen2024knowledge}$_\text{ECCV'2024}$    & 67.0 & 73.8 \\

        \midrule
        CAMs                                                & 62.5 & -    \\
        CAMs $+$ Affinity                                   & 65.9 (+3.4) & -    \\
        CAMs $+$ Affinity $+$ AReAM                         & 68.9 (+6.4) & -    \\
        CAMs $+$ Affinity $+$ AReAM $+$ PAR~\cite{2022AFA}  & \textbf{69.2} (+6.7) & \textbf{75.1} \\
        \bottomrule
    \end{tabular}
\end{table}

\begin{table}[htbp]
    \centering
    \caption{Ablation results of the weight coefficient $\lambda$.}
    \label{tab:lambda}
    \begin{tabular}{c|cccccc}
        \toprule
        $\lambda$ & 0.1  & \textbf{0.2}  & \underline{0.3}  & \textbf{0.4}  & 0.5  \\
        \midrule
        mIoU (\%) & 67.6 & \textbf{69.2} & \underline{69.1} & \textbf{69.2} & 68.8 \\
        \bottomrule
    \end{tabular}
\end{table}

\noindent \textbf{Ablation of AReAM.}
Table~\ref{tab:pseudo_label_result} reports the semantic segmentation performance of pseudo labels $\mathbf{Y}_s$ on the PASCAL VOC training set. As in \cite{2020SEAM,2021AdvCAM,2022MCTformer}, we present the optimal results by testing a range of background thresholds. Notably, AReAM boosts the quality of pseudo labels both with and without post-processing, improving performance from $62.5\%$ to $69.2\%$ by refining CAMs and reaching $75.1\%$ after further post-processing. In particular, AReAM outperforms the baseline, where CAMs are refined using the average affinity matrix $\bar{\mathbf{M}}$, by approximately $3.4\%$. Although incorporating PAR~\cite{2022AFA} yields an additional increase of $6.7\%$, the direct improvement from AReAM itself is $6.4\%$, demonstrating that mitigating over-smoothing through calibrated attention relations is both necessary and effective.


\noindent \textbf{Layer-wise Gains of AReAM.}
We validate the layer-wise gains achieved by AReAM on PASCAL VOC 2012 for vision transformers with 6, 8, 10, and 12 layers. As depicted in Fig.~\ref{fig:layerwise_gain}, the blue and orange horizontal lines represent the performance of refining CAMs without and with AReAM, respectively. Dashed lines with markers show layer-wise outcomes, and the orange shaded region highlights the improvements from our re-activation mechanism. Notably, AReAM effectively alleviates the performance degradation observed at deeper layers; for instance, it improves the last layer of a 6-layer vision transformer by approximately 20\%. Interestingly, shallow-layer performance also benefits, which can be attributed to the counterpart weights allowing shallow layers to participate in re-activation alongside deeper layers.

\begin{figure}[!htb]
\centering
    \includegraphics[width=\linewidth]{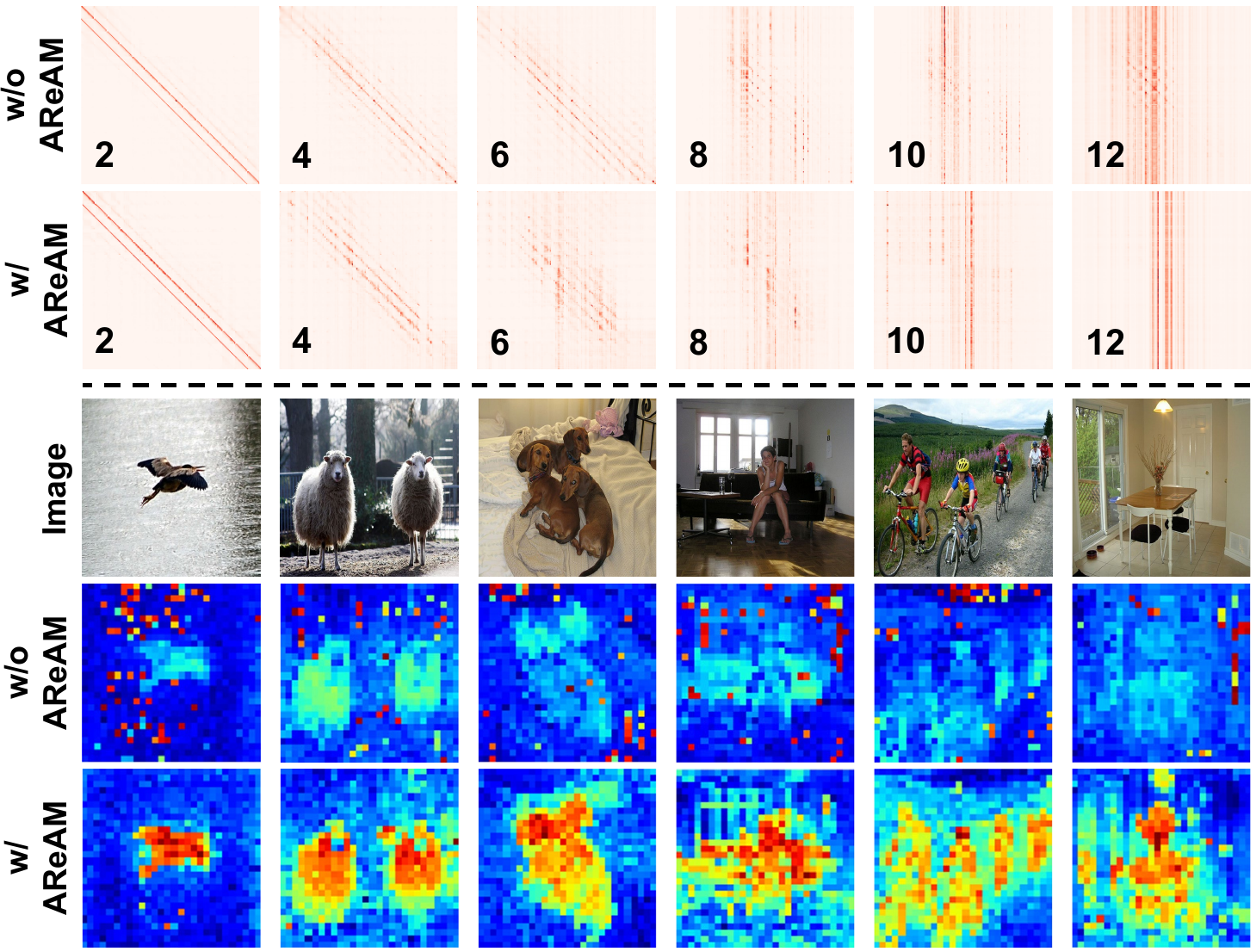}
    \caption{(top) Visualization of the layer-wise affinity matrices. (bottom) Visualization of the attended key tokens. We take the average of the last-layer affinity matrix along the query dimension, and reshape the 1-dimensional sequence to a 2-dimensional heatmap.}
    \label{fig:affinity_matrix_comparison}
\end{figure}

\begin{figure*}[t]
    \centering
    \includegraphics[width=\textwidth]{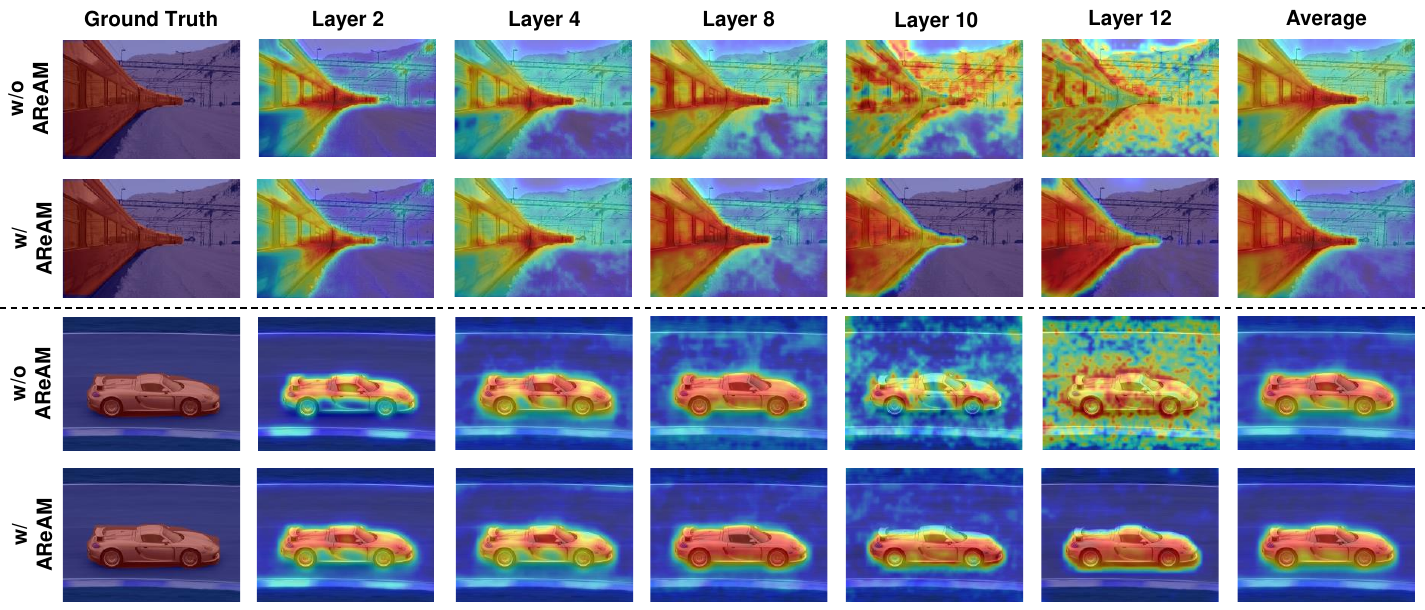}
    \caption{\small Visualization of CAMs refined by the affinity matrix of each layer, and the average affinity matrix (the last column).}
    \label{fig:layerwise_visualization}
\end{figure*}

\noindent \textbf{Analysis of Undisciplined Over-Smoothing.}
In Fig.~\ref{fig:affinity_matrix_comparison} (top), we illustrate that over-smoothing is an intrinsic property of vision transformers. Regardless of whether AReAM is applied, each query token predominantly attends to its own key token at shallow layers (diagonal pattern), and converges to a similar subset of key tokens at deeper layers (strip pattern). However, at sufficiently deep layers (e.g., the 12\textsuperscript{th} layer), the specific tokens receiving high attention differ substantially once AReAM is enabled.

In Fig.~\ref{fig:affinity_matrix_comparison} (bottom), we examine how AReAM alters the attended tokens by averaging the last-layer affinity matrix across all query tokens and reshaping the resulting 1D sequence into a 2D heatmap for visualization. Strikingly, when trained without re-activation (rows labeled “w/o AReAM”), the vision transformer focuses predominantly on background tokens—consistent with earlier findings~\cite{darcet2023vision} that well-trained transformers rely on redundant tokens to gather global information. In contrast, AReAM directs attention toward semantically meaningful object regions (rows labeled “w/ AReAM”), effectively mitigating the \emph{undisciplined over-smoothing} that hampers WSSS performance.

Finally, Fig.~\ref{fig:layerwise_visualization} provides layer-wise examples of CAMs refined by the corresponding affinity matrices. Without re-activation, deeper layers devote excessive attention to background areas, resulting in significant noise (top row of each example). By contrast, AReAM adaptively corrects the affinities at every layer, yielding refined CAMs that remain largely free of background noise (bottom row of each example).

\begin{table}[htbp]
    \centering
    \setlength{\tabcolsep}{2pt}
    \caption{Effect of directly fusing the first $K$ layers. The results of pseudo labels are reported on PASCAL VOC 2012 in terms of mIoU (\%). The highest results are in bold; the second highest results are with underline.}
    \resizebox{\linewidth}{!}{
    \begin{tabular}{l|cccccccccccc}
    \toprule
    $K$ & 1 & 2 & 3 & 4 & 5 & 6 & 7 & 8 & 9 & 10 & 11 & 12 \\
    \midrule
    AReAM ($\times$)     & 51.7 & 63.5 & 64.3 & 64.5 & 65.0 & 65.3 & \textbf{66.1} & \underline{66.0} & \textbf{66.1} & \textbf{66.1} & \textbf{66.1} & 65.9 \\
    AReAM ($\checkmark$) & 56.7 & 64.1 & 64.6 & 64.8 & 65.4 & 66.0 & 67.0 & 67.4 & 68.2 & 68.6 & \underline{69.1} & \textbf{69.2} \\
    \bottomrule
    \end{tabular}}
    \label{tab:shallow_layer_ablation}
\end{table}

\begin{figure*}[!htb]
    \centering
    \includegraphics[width=0.95\textwidth]{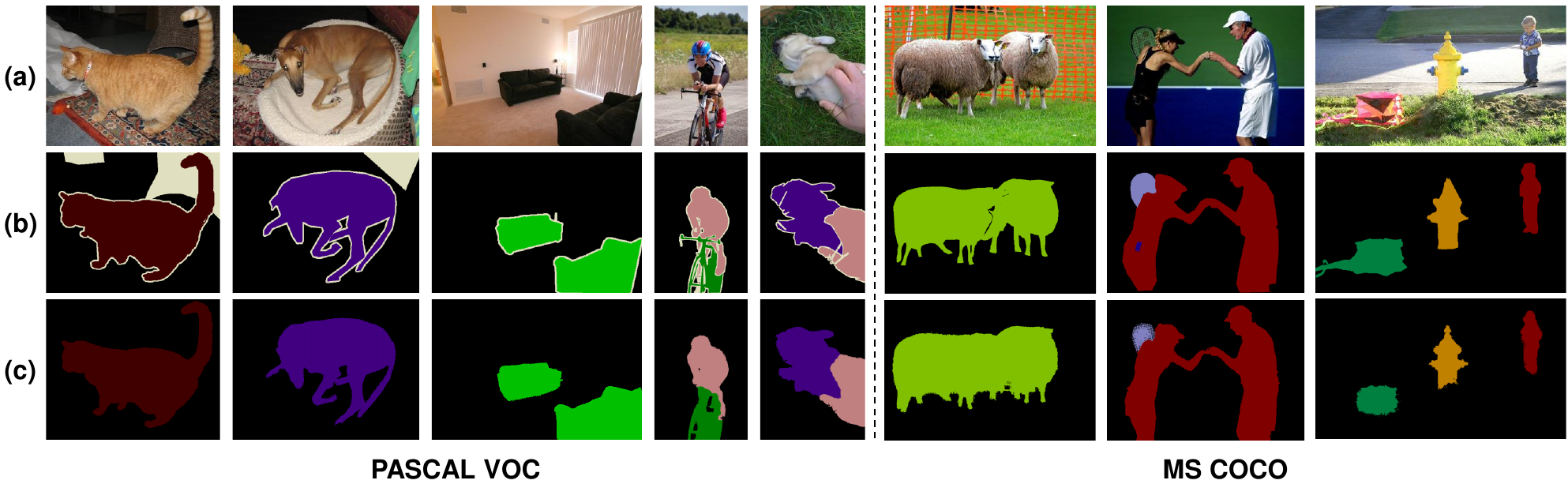}
    \caption{Visualization of semantic segmentation results on PASCAL VOC and MS COCO \textit{val} sets. (a) Image. (b) Ground truth. (c) Ours.}
    \label{fig:qualitative_results}
\end{figure*}

\noindent \textbf{Can We Directly Fuse Shallow Affinity Matrices?}
Although shallow layers contain less noise, they mainly capture low-level and localized features, making them insufficient for recognizing entire objects. As shown in Fig.~\ref{fig:layerwise_visualization}, the second layer, for instance, fails to detect the left portion of the ``train.'' Table~\ref{tab:shallow_layer_ablation} provides a quantitative analysis of directly fusing shallow-layer affinities. When AReAM is disabled, performance increases only until fusing up to layer~7, suggesting limited gains when deeper layers (8--12) are also integrated. In contrast, when AReAM is enabled, performance steadily improves as more layers are fused, and using all layers achieves the best result.

\noindent \textbf{Ablation of Hyper-parameters.}
As discussed in Section~\ref{sec:method}, the re-activation and counterpart weights are adaptively adjusted according to the entropy of the activations. Consequently, the only additional hyper-parameter introduced by AReAM is the loss coefficient $\lambda$ in Eq.~(\ref{eq:total_loss}). Table~\ref{tab:lambda} shows the ablation study on $\lambda$. We empirically set $\lambda = 0.2$, which balances classification and re-activation effectively.

\subsection{Comparison to State-of-the-arts}
\label{sec:SOTA}
Following the common practice of WSSS methods, we further demonstrate the effectiveness of AReAM by training a semantic segmentation model with our generated pseudo labels. The comparison results with previous state-of-the-arts on the PASCAL VOC and MS COCO datasets are displayed in Tab. \ref{tab:sem_seg}. It shows that our method consistently surpasses previous image-level WSSS methods on both benchmarks. 
Specifically, our approach using Deeplab-V1 with a ResNet38
backbone (``V1-RN38'' in Tab. \ref{tab:sem_seg})~\cite{chen2014semantic} achieves 73.7\% and 73.4\% on the \textit{val} and \textit{test} sets of PASCAL VOC, respectively, and 44.8\% on the \textit{val} set of MS COCO. 
These figures are slightly higher than those obtained using Deeplab-V2 with a ResNet101 backbone (``V2-RN101'' in Tab. \ref{tab:sem_seg})~\cite{deeplab}. 
We also provide a few examples of predicted semantic segmentation labels in Fig. \ref{fig:qualitative_results}, which shows visually plausible results compared to the ground truth. 

\begin{table}[t]
    \centering
    \caption{Semantic segmentation results in terms of mIoU (\%) on PASCAL VOC and MS COCO. ``V1'' and ``V2'' refer to DeepLab-V1 and DeepLab-V2~\cite{deeplab} respectively, ``RN'' means ResNet~\cite{he2016deep,wu2019wider}.}
    \label{tab:sem_seg}
    \begin{tabular}{l l | c c | c}
        \toprule
        \multirow{2}{*}{Method} & \multirow{2}{*}{Network}  & \multicolumn{2}{c|}{VOC($\uparrow$)}  & COCO($\uparrow$) \\
                                &                           & \textit{val} & \textit{test}          & \textit{val}   \\
        \midrule
        PSA~\cite{2018PSA}$_{\text{CVPR'2018}}$             & V1-RN38 & 61.7 & 63.7 & -    \\
        IRN~\cite{2019IRN}$_{\text{CVPR'2019}}$             & V2-RN50 & 63.5 & 64.8 & -    \\
        SEAM~\cite{2020SEAM}$_{\text{CVPR'2020}}$           & V1-RN38 & 64.5 & 65.7 & -    \\
        CONTA~\cite{zhang2020causal}$_{\text{NeurIPS’2020}}$& V1-RN38 & 66.1 & 66.7 & 32.8 \\
        BES~\cite{2020BES}$_{\text{ECCV'2020}}$             & V2-RN101& 65.7 & 66.6 & -    \\
        CDA~\cite{su2021context}$_{\text{ICCV'2021}}$       & V1-RN38 & 66.1 & 66.8 & 33.2 \\
        ECS-Net~\cite{sun2021ecs}$_{\text{ICCV'2021}}$      & V1-RN38 & 66.6 & 67.6 & -    \\
        CPN~\cite{2021CPN}$_{\text{ICCV'2021}}$             & V1-RN38 & 67.8 & 68.5 & -    \\
        AdvCAM~\cite{2021AdvCAM}$_{\text{CVPR'2021}}$       & V2-RN101& 68.1 & 68.0 & 44.4 \\
        RIB~\cite{2021RIB}$_{\text{NeurIPS'2021}}$          & V2-RN101& 68.3 & 68.6 & 43.8 \\
        AFA~\cite{2022AFA}$_\text{CVPR'2022}$               & MiT-B1  & 66.0 & 66.3 & 38.9 \\
        W-OoD~\cite{lee2022weakly}$_\text{CVPR'2022}$       & V1-RN38 & 70.7 & 70.1 & -    \\
        AMR~\cite{2022AMR}$_{\text{AAAI'2022}}$             & V2-RN101& 68.8 & 69.1 & -    \\
        ESOL~\cite{li2022expansion}$_\text{NeurIPS’2022}$   & V2-RN101& 69.9 & 69.3 & 42.6 \\
        AMN~\cite{lee2022threshold}$_\text{CVPR'2022}$      & V2-RN101& 70.7 & 70.6 & 44.7 \\
        MCTformer~\cite{2022MCTformer}$_{\text{CVPR'2022}}$ & V1-RN38 & 71.9 & 71.6 & 42.0 \\
        FPR~\cite{chen2023fpr}$_\text{ICCV'2023}$           & V2-RN101& 70.3 & 70.1 & 43.9 \\
        ToCo~\cite{ru2023token} $_\text{CVPR'2023}$         & ViT-B   & 71.1 & 72.2 & 42.3 \\
        USAGE~\cite{jo2023mars}$_\text{ICCV'2023}$          & V1-RN38 & 71.9 & 72.8 & 42.7 \\
        LPCAM~\cite{chen2023extracting}$_\text{CVPR'2023}$  & V1-RN38 & 72.6 & 72.4 & 42.8 \\
        ACR~\cite{kweon2023weakly}$_\text{CVPR'2023}$       & V1-RN38 & 72.4 & 72.4 & -    \\
        OCR~\cite{cheng2023out}$_\text{CVPR'2023}$          & V1-RN38 & 72.7 & 72.0 & 42.5 \\
        EPG~\cite{qin2024enhanced}$_\text{TCSVT'2024}$          & V2-RN50  & 67.3 & 67.5 & - \\
        MCC~\cite{Wu_2024_WACV}$_\text{WACV'2024}$          & DeiT-B  & 70.3 & 71.2 & 42.3 \\
        CPRE~\cite{su2024cross}$_\text{IJCNN'2024}$         & V2-RN101  & 72.5 & 71.9 & 44.8 \\
        DuPL~\cite{wu2024dupl}$_\text{CVPR'2024}$           & ViT-B   & 73.3 & 72.8 & 44.6 \\
        
        \midrule
        Ours                                                & V2-RN101& 73.3 & 72.9 & 44.8 \\
        Ours                                                & V1-RN38 & \textbf{73.7} & \textbf{73.4} & \textbf{45.1} \\
        \bottomrule
    \end{tabular}
\end{table}

\section{Conclusion}
\noindent This study inspects the undisciplined over-smoothing issue that impairs WSSS performance, and introduces an Adaptive Re-Activation Mechanism (AReAM) as a solution to addressing it. AReAM exploits shallow-level affinities to instruct deep-level tokens to converge towards semantic objects. 
The effectiveness of our proposed AReAM is evidenced by extensive experiments conducted on established public datasets. 
Furthermore, it holds promise for applications in various transformer-based re-activation tasks. 


\bibliographystyle{unsrt}
\bibliography{main}

\end{document}